\pdfoutput=1

\documentclass[11pt]{article}

\usepackage{acl}
\usepackage{times}
\usepackage{latexsym}
\usepackage[T1]{fontenc}
\usepackage[utf8]{inputenc}
\usepackage{microtype}
\usepackage{inconsolata}
\usepackage{hyperref}
\usepackage{url}
\usepackage{booktabs}  
\usepackage{amsfonts}
\usepackage{amsmath}
\usepackage{amssymb}
\usepackage{graphicx}  
\usepackage{multirow}
\usepackage{multicol}
\usepackage{xcolor}    
\usepackage{pifont}    
\usepackage{algorithm}
\usepackage{algorithmic}
\usepackage{siunitx}
\usepackage{float}     
\usepackage{colortbl}

\graphicspath{{images/}}

\definecolor{darkgreen}{rgb}{0.0, 0.5, 0.0}
\definecolor{lightgreen}{RGB}{120,200,120}
\definecolor{lightred}{RGB}{230,120,120}
\definecolor{methodbg}{rgb}{0.93, 0.88, 0.98}
\definecolor{alggreen}{rgb}{0.0, 0.45, 0.3}

\newcommand{\ours}{\textsc{CACD}}

\newcommand{\mask}{[\text{MASK}]}

\newcommand{\algcomment}[1]{\textcolor{alggreen}{\small\textit{// #1}}}

\title{Context-Aware Cluster Decoding: Semantic Anchor-Driven\\
Coherence in dMLLMs}

\author{
Yikai Zhao\textsuperscript{2}\thanks{This work is supported by Open Research Fund of The State Key Laboratory of Multimodal Artificial Intelligence Systems (MAIS2026079).}, Qiyan Zhao\textsuperscript{3}, 
Jiaquan Zhang\textsuperscript{4}, Xiaofeng Zhang\textsuperscript{3} \\
\textbf{Xiaosong Yuan}\textsuperscript{5}, 
\textbf{Pengzhou Cheng\textsuperscript{1}}\thanks{\ \ Corresponding author.}
\\
\textsuperscript{1}Shanghai University \quad
\textsuperscript{2}Sun Yat-sen University \quad
\textsuperscript{3}Shanghai Jiao Tong University
\\
\textsuperscript{4}University of Electronic Science and Technology of China \quad
\textsuperscript{5}Alibaba Group 
\\
\texttt{zhaoyk25@mail2.sysu.edu.cn, chengpz@shu.edu.cn}
}

\begin{document}
\maketitle

\begin{abstract}
Diffusion multimodal large language models (dMLLMs) frequently
produce long-form outputs marred by semantic drift and repetition,
with quality generally degrading as output length increases.
We identify two structural deficiencies in existing decoding methods
as primary drivers of these failures: confidence-based scoring
ignores decoded-neighbor support, and block partitioning prevents
access to high-readiness semantic anchors, together causing tokens
to be committed before their local context is sufficiently
established.
We propose \ours{} (\textbf{C}ontext-\textbf{A}ware \textbf{C}luster
\textbf{D}ecoding), a training-free decoding method that scores each
masked position by a multiplicative composite of softmax confidence
and neighbor proximity, promoting contextually ready tokens above
isolated candidates while suppressing low-confidence positional
noise, operating block-free to keep high-readiness anchors
globally accessible. \ours{} further applies architecture-aware
calibration to handle confidence heterogeneity induced by diverse
visual integration strategies.
Experiments on three dMLLMs across four benchmarks demonstrate
consistent quality gains and hallucination reduction over Original,
with larger gains in several longer generation settings, highlighting
the importance of neighbor support and visual integration strategy
for future dMLLM decoding method design. Our code is openly available
at \url{https://github.com/zhaoyk-sysu/CACD-dMLLM}.
\end{abstract}

\section{Introduction}
\label{sec:intro}
\begin{figure}[t]
  \centering
  \includegraphics[width=\columnwidth]{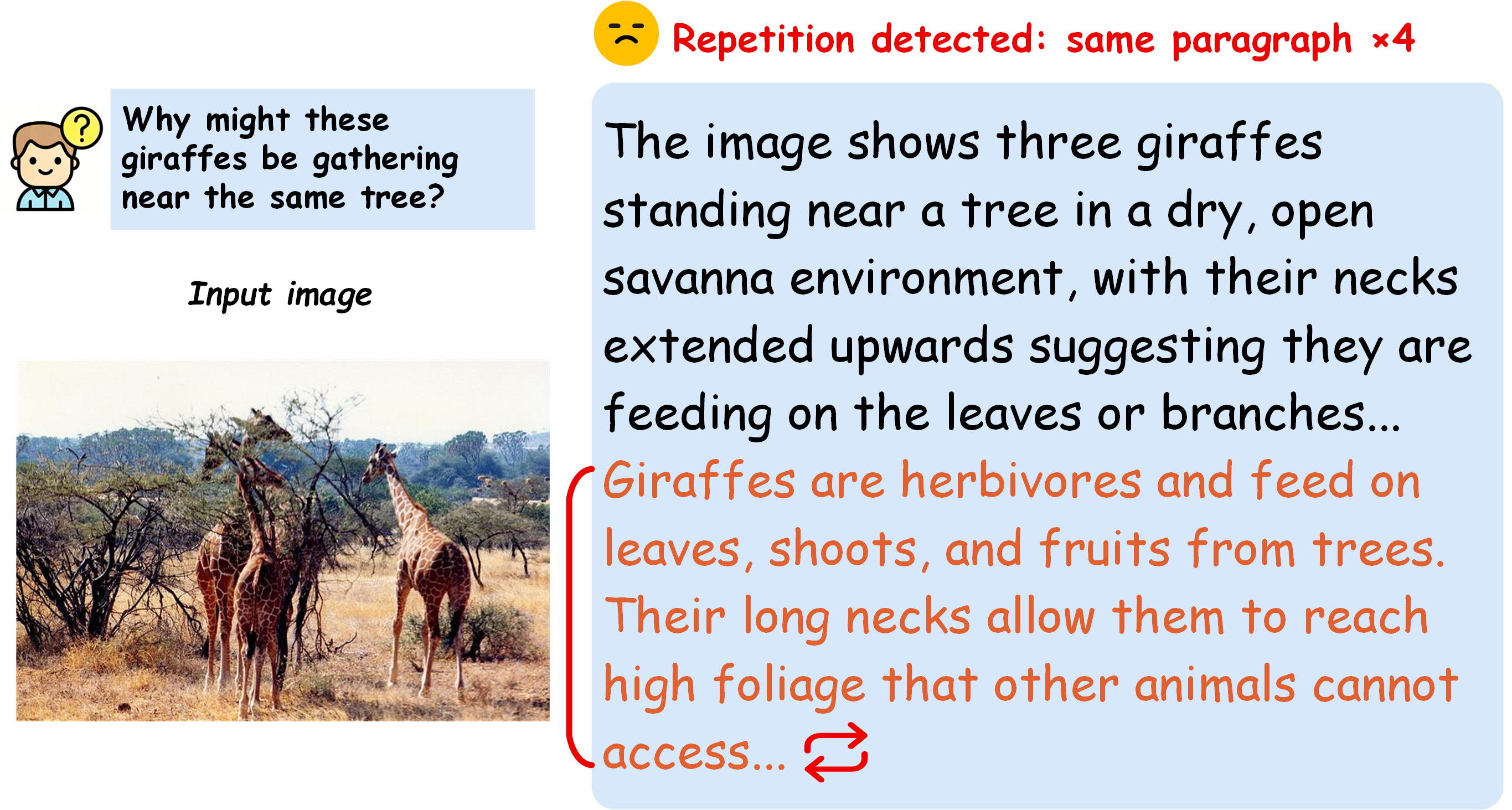}
   \vspace{-4pt}
  \caption{A representative block-diffusion failure: the same paragraph
    repeats four times, a symptom of incorrect commitment order leaving
    neighboring positions without local semantic support.}
    \vspace{-3pt}
  \label{fig:motivation}
\end{figure}

\begin{figure*}[t]
  \centering
  \includegraphics[width=\textwidth]{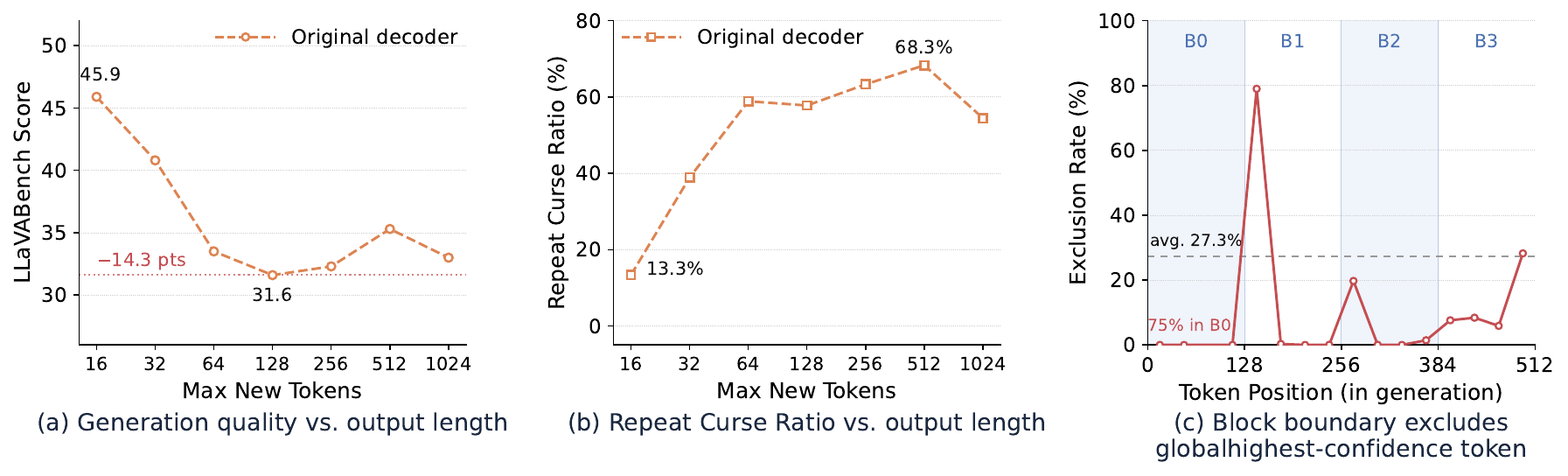}
  \vspace{-6pt}
  \caption{Two failure modes of the Original block-diffusion method
  on MMaDA-8B-Base.
  \textbf{(a)} LLaVABench-COCO score degrades monotonically with
  output length.
  \textbf{(b)} Sample Repeat Curse Ratio on LLaVABench-COCO increases
  with output length, measured following~\cite{cota}.
  \textbf{(c)} A high-readiness semantic anchor candidate lies outside
  the current block in \textbf{75.0\%} of first-round steps and
  \textbf{27.3\%} of all steps (block\_length$=128$, 90
  LLaVABench-COCO samples), revealing that block partitioning systematically prevents access to highly effective anchors.}
  \vspace{-6pt}
  \label{fig:motivation_failure}
\end{figure*}

Diffusion multimodal large language models
(dMLLMs~\cite{llada,llada2,mmada,lavida,lladav,dimple,unified})
have emerged as a compelling alternative to autoregressive
generation, offering parallel decoding under bidirectional
attention as a principled departure from left-to-right token
commitment. Despite this architectural advantage, dMLLMs
frequently produce long-form outputs marred by semantic drift
and repetitive phrasing (Figure~\ref{fig:motivation}), with
quality gaps widening monotonically with output length
(Figure~\ref{fig:motivation_failure}).

\begin{figure*}[t]
  \centering
  \includegraphics[width=\textwidth]{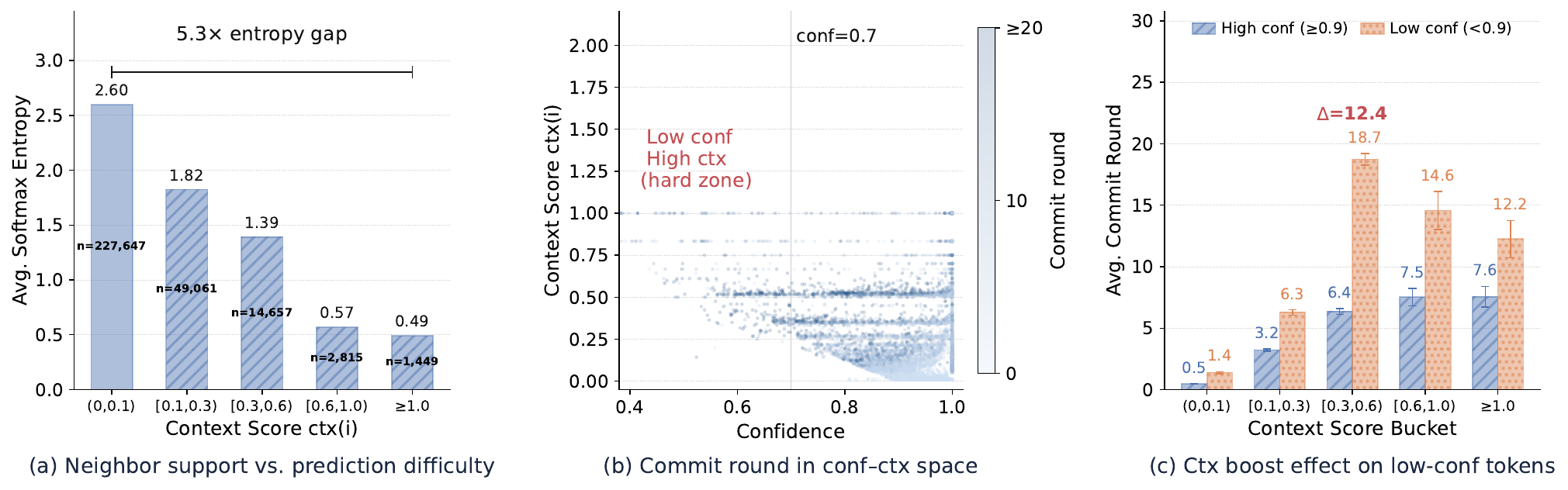}
  \vspace{-4pt}
  \caption{Motivation analysis (MMaDA, 1,000 MathVista samples).
  \textbf{(a)} Neighbor support vs.\ prediction difficulty: entropy
  drops $5.3\times$ from isolated to fully-neighbored
  positions---is strongly associated with prediction difficulty beyond raw confidence
  \textbf{(b)} Confidence--context space colored by commit round;
  the ``hard zone'' marks tokens with low confidence but high context
  score that are contextually ready yet stranded by pure confidence
  scoring.
  \textbf{(c)} Commit-round gap peaking at $\Delta{=}12.4$ rounds,
  quantifying how long voids persist under pure confidence scoring.}
  \vspace{-4pt}
  \label{fig:motivation_analysis}
\end{figure*}

Our analysis suggests that one important source of these
degradations lies in how existing decoders handle commitment
order: they neither identify high-quality semantic anchors nor
exploit local semantic connections between decoded tokens and
their masked neighbors, which causes positions to be committed
in an order that leaves subsequent tokens without sufficient
local support.
The key insight is that local connections to decoded
neighbors substantially affect contextual readiness beyond 
raw confidence: on 1000 MathVista samples, positions with both
neighbors decoded achieve $5.3\times$ lower posterior entropy
than isolated positions, suggesting that decoded neighbors
provide strong local semantic
constraint~\cite{kim2025} (Figure~\ref{fig:motivation_analysis}a).
When a contextually ready token is wrongly deferred, a weaker
candidate is committed in its place, entering surrounding context
without sufficient local support and degrading subsequent
predictions. This effect compounds with sequence length,
ultimately producing semantic drift and hallucination.

Two structural obstacles prevent existing decoding methods from exploiting these properties.

\textbf{Finding 1: Confidence ignores neighbor support.}
Existing scoring methods~\cite{fastdllm,apd,ccd} rank positions by
raw confidence, entropy, or autoregressive mixtures, none of which
measures the local semantic support a position's neighborhood already
provides.
Tokens that are contextually ready (strongly anchored by committed
neighbors) are systematically ranked below isolated tokens of
comparable confidence, causing the latter to be committed first
despite weaker local support; conversely, low-confidence tokens at
favorable positions incur no penalty under pure confidence scoring.

\textbf{Finding 2: Block structure can exclude high-readiness anchor candidates.}
Even with accurate scoring, block-based
decoding methods~\cite{bd3,adablock,dcd} cannot always access the
highest-readiness anchor candidate.
This exclusion is most damaging during the first decoding round, when
committed tokens establish the primary semantic anchors for all
subsequent positions: \textbf{75.0\%} of first-round steps find the
globally highest-readiness anchor outside the current block
(Figure~\ref{fig:motivation_failure}b), and the exclusion persists in
\textbf{27.3\%} of all steps, meaning over one in four opportunities
to commit the strongest available anchor is structurally blocked.
AdaBlock~\cite{adablock} and DCD~\cite{dcd} progressively reduce
this misalignment but cannot fully eliminate it without abandoning
sequential block structure.

Both findings deny subsequent positions the local semantic support
needed for reliable prediction. We propose \ours{}
(\textbf{C}ontext-\textbf{A}ware \textbf{C}luster \textbf{D}ecoding),
which scores every masked position by a multiplicative composite of
softmax confidence and neighbor proximity, operates block-free to
keep the most informative anchor always accessible, and applies
architecture-aware calibration to handle confidence heterogeneity
across dMLLM architectures. Extensive experiments across three
dMLLMs and four benchmarks confirm that \ours{} provides an
effective and broadly applicable commitment signal.

Our contributions are:
\begin{itemize}
\item We identify that existing decoding methods systematically fail
to satisfy two critical properties: semantic anchor accessibility and
contextual readiness ordering. We show this failure significantly
limits long-form generation quality in dMLLMs, providing empirical
evidence that block partitioning cuts off high-readiness anchors in
\textbf{75.0\%} of first-round steps and \textbf{27.3\%} of all
steps.
\item We introduce \ours{}, a plug-and-play decoding framework for dMLLMs. 
We demonstrate that a multiplicative composite score of softmax confidence 
and local semantic context simultaneously rescues contextually ready tokens 
and suppresses positional noise. This joint formulation provides a more 
effective commitment signal than confidence alone while maintaining global 
anchor accessibility.
\item We show that distinct visual integration strategies shape the
confidence landscape in ways existing text-only decoding methods do
not account for, a dimension that serves as an essential consideration 
for future dMLLM decoding method design.
\end{itemize}

\section{Related Work}
\label{sec:related}

The masked diffusion decoding process~\cite{llada} starts from a fully masked sequence of tokens and iteratively unmasks positions, where at each step the model predicts all masked tokens in parallel and a sampler selects a subset to commit based on confidence or other criteria.

\paragraph{Commitment Order and Prediction Difficulty.}
Kim et al.~\cite{kim2025} establish that commitment order is a
primary determinant of generation quality, and that committing
high-certainty positions first substantially outperforms fixed-order
decoding. Zhou et al.\ proposed HDLM~\cite{hdlm} (NeurIPS~2025),
which reinforces this from the training perspective via a
coarse-to-fine semantic hierarchy. Neither work provides an
inference-time mechanism that maintains correct commitment order
under complex visual context or long-form outputs, leaving the
resulting quality cost unaddressed;
Section~\ref{sec:mot_semantic} quantifies it directly.

\paragraph{Decoding Order and Block Structure.}
Arriola et al.\ proposed Block Diffusion~\cite{bd3}, partitioning
sequences into fixed blocks. AdaBlock~\cite{adablock} (ICLR~2026),
DCD~\cite{dcd} and others
\cite{zhang1,zhang2,zhang3,zhang4,zhang5,zhang6,zhang7,zhang8,zhang9}
refine block boundaries in a more flexible manner but retain strict
sequential structure, structurally preventing access to
highest-readiness anchor candidates.
WavefrontDiffusion~\cite{wavefront} (ICLR~2026) expands a frontier
from finalized positions and explicitly models spatial neighbor
distance, yet its commitment signal remains confidence-based without
incorporating bilateral neighbor support into scoring.
AHD~\cite{ahd} enables early cross-block decoding via historical
convergence trends, requiring a multi-step history buffer; \ours{}
achieves neighbor-aware commitment within a single forward pass.
Kang et al.~\cite{parallelbench} confirm that token dependency
strength is the primary determinant of decoding quality, directly
supporting neighborhood support as the key commitment signal.

\paragraph{Confidence-Based Scoring.}
Fast-dLLM~\cite{fastdllm}, APD~\cite{apd}, and SlowFast
Sampling~\cite{slowfast} (ICLR~2026) improve commitment through
confidence-threshold and dynamic scoring strategies~\cite{conftheory},
but treat confidence as the sole signal without measuring neighborhood support.
Chen et al.\ proposed Coherent Contextual Decoding~\citep{ccd},
which identifies semantically pivotal tokens via cross-step
consistency in a sliding history buffer and prioritizes their
commitment. By contrast, our context score operates purely in the
spatial domain, measuring proximity to currently decoded bilateral
neighbors within a single forward pass; the two approaches thus
capture fundamentally different notions of context, temporal
consistency versus spatial neighborhood support, and are complementary.
Prophet~\cite{prophet} identifies a two-phase commitment risk pattern
that motivates our dynamic threshold scheduling; further work
addresses scoring pathologies~\cite{uncode,remedi} and repetition
caused by caching~\cite{cota}. Hong et al.~\cite{mpdpac2026} approach
the same symptoms from the attention level, proposing training-free
interventions to address repetition and visual grounding degradation;
by contrast, \ours{} operates at the commitment-order level without
modifying attention mechanisms.

\paragraph{Summary.}
Existing anchor-seeking methods either rely on the multi-step temporal history buffers of AHD~\cite{ahd} for contextual consistency, or restrict the decoding frontier via architectural spatial distance constraints as in WavefrontDiffusion~\cite{wavefront}. 
Crucially, neither translates spatial proximity into a dynamic scoring signal, leaving bilateral neighbor support unmeasured during the commitment phase. 
Furthermore, block-based frameworks structurally isolate high-readiness candidates across boundaries, and confidence heterogeneity from multimodal integration remains unaddressed. 
By contrast, \ours{} introduces a global, single-pass bilateral neighbor proximity score, transforming structural constraints into a direct commitment signal. 
A broader discussion is provided in Appendix~\ref{app:related}.

\section{Motivation}
\label{sec:motivation}

\subsection{Local Semantic Connections Determine Prediction Quality}
\label{sec:mot_semantic}

Let $\mathbf{x} = [x_1, \ldots, x_L]$ be the generation sequence,
where position $i$ is either a decoded token or $\mask$.
In dMLLMs, visual inputs are encoded and prepended as conditioning
context, with integration strategies varying across architectures.
At each step the model performs a full-sequence forward pass for
logits $\mathbf{l} \in \mathbb{R}^{L \times V}$; the confidence of
position $i$ is $c_i =
\max_{v}\operatorname{softmax}(\mathbf{l}_i)_v$. A sampler selects
a subset of masked positions to commit, repeating until all positions
are decoded.

We define \textit{contextual readiness} as the strength of local
semantic connections a position shares with its already-decoded
bilateral neighborhood.
Figure~\ref{fig:motivation_analysis}a quantifies this over 295,629
masked positions: positions with both neighbors decoded achieve
average entropy $0.49$ versus $2.60$ for isolated positions, a
$\mathbf{5.3\times}$ gap confirming that decoded neighbors provide
strong additional signal beyond raw confidence~\cite{kim2025}.

\paragraph{From observation to design.}
This entropy gap directly motivates the composite score: a token with
both neighbors decoded is far more predictable than an isolated one,
yet pure confidence scoring treats them identically, committing the
isolated token first if its confidence is higher.
The composite score corrects this by incorporating neighbor proximity
as a multiplicative amplifier on confidence, reflecting the measured
entropy advantage.

\paragraph{The consequence of ignoring neighbor support.}
Pure confidence scoring strands contextually ready tokens below
threshold despite having both neighbors decoded.
These tokens persist as $\mask$ voids: adjacent positions must
predict across the gap using weaker, more distant context.
We term this cascading effect \textit{void compounding}: each
misdeferred token creates a void that weakens local support for its
neighbors, propagating further misdeferral monotonically with
sequence length.
As shown in Figure~\ref{fig:motivation_analysis}b, tokens with low
confidence but high context score are systematically stranded in the
``hard zone'' under pure confidence scoring.
Figure~\ref{fig:motivation_analysis}c quantifies the cost directly:
the commit-round gap peaks at $\Delta{=}12.4$ rounds, confirming
that voids persist long enough to substantially degrade local
semantic support for neighboring positions. During this interval,
surrounding positions must predict across an expanding gap, driving
their context scores toward zero and effectively reducing their
scoring to pure confidence, the same regime that caused the
original misdeferral.
A step-by-step visualization is provided in
Appendix~\ref{app:decoding_order} (Figure~\ref{fig:decoding_comparison}).

\begin{figure*}[t]
  \centering
  \includegraphics[width=\textwidth]{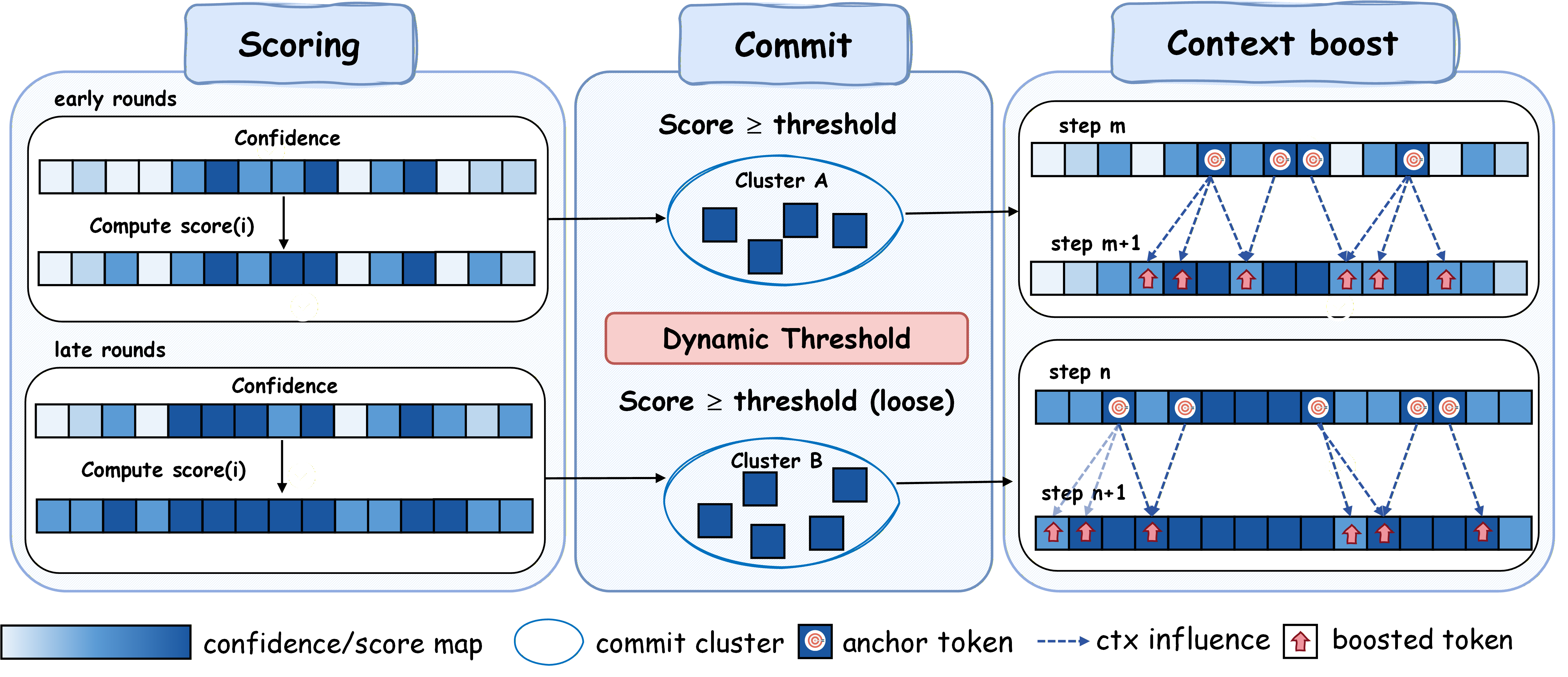}
  \vspace{-4pt}
  \caption{Overview of \ours{}.
    \textit{Left}: the Scoring phase computes a composite score for
    each masked position across early and late rounds; early rounds apply a
    strict threshold $\tau$, while late rounds apply a relaxed
    dynamic threshold triggered when $r_t \leq \delta$.
    \textit{Center}: positions with composite scores exceeding the current threshold 
    are committed simultaneously as a dense cluster.
    \textit{Right}: committed anchor tokens propagate localized context
    influence (dashed lines) to adjacent masked positions, boosting their scores for the subsequent decoding round.}
    \vspace{-6pt}
  \label{fig:comparison}
\end{figure*}

\subsection{Block Structure Can Exclude Globally Preferred Anchor Candidates}
\label{sec:mot_structural}
Even with accurate scoring, block-based methods may fail to access
high-readiness anchor tokens, whose exclusion leaves neighboring
positions without strong local support.
This is most damaging during the first decoding round, when committed
tokens become the primary semantic anchors for all subsequent
positions, as quantified in Figure~\ref{fig:motivation_failure}c.
Increasing block size cannot resolve this, as any fixed sequential
partition creates boundaries where high-readiness anchors may be
unavailable.

The two failure modes compound: block exclusion forces weaker tokens
to be committed first, and each such commit creates a void that
propagates through subsequent positions via the same void-compounding
mechanism described in Section~\ref{sec:mot_semantic} --- meaning
degradation begins from the very first decoding round rather than
accumulating gradually.
Appendix~\ref{app:decoding_order} further illustrates how block
boundaries delay high-readiness tokens despite favorable scores.
\ours{} addresses both with a single unified approach, described in
Section~\ref{sec:method}.

\section{Method}
\label{sec:method}

\ours{} is built on a single principle: generation quality in dMLLMs
depends on identifying high-quality semantic anchors and exploiting
local semantic connections between decoded tokens and their masked
neighbors.
Two components instantiate this principle: a composite score
(Section~\ref{sec:scoring}) and our cluster-based commitment approach
(Section~\ref{sec:global}).
Three calibration components ensure stable operation across
architectures and output lengths (Section~\ref{sec:calibration}).
Figure~\ref{fig:comparison} illustrates the three-phase pipeline of \ours{}.

\subsection{Context-Aware Scoring}
\label{sec:scoring}

To ensure contextually ready tokens are committed before isolated
tokens of comparable confidence, \ours{} incorporates neighbor
proximity directly into the commitment score.

\paragraph{Token Confidence.}
\begin{equation}
  c_i = \max_{v \in \mathcal{V}}\operatorname{softmax}(\mathbf{l}_i)_v.
  \label{eq:conf_def}
\end{equation}

\begin{figure*}[t]
  \centering
  \includegraphics[width=\textwidth]{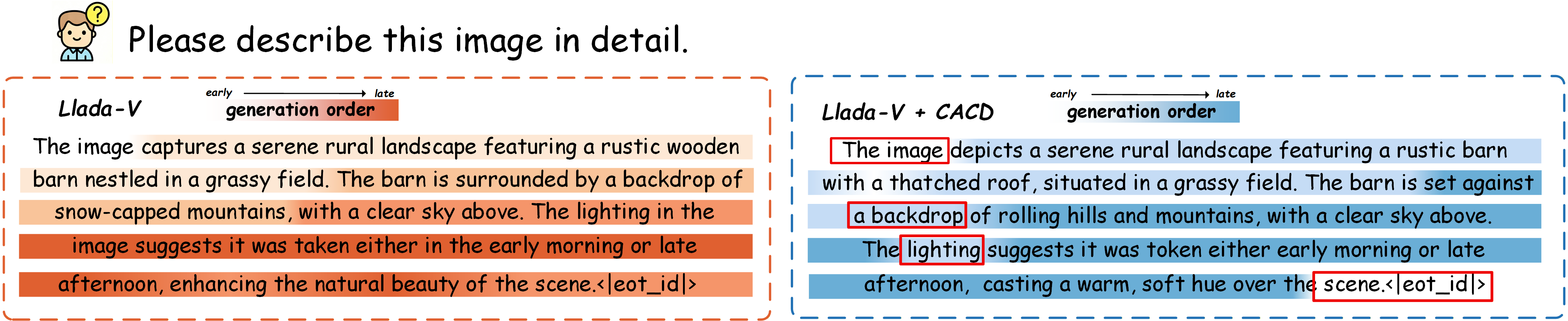}
  \vspace{-4pt}
  \caption{Decoding order on a representative captioning sample
    (lighter = earlier commit; red boxes = semantic anchors).
    Block diffusion frequently bypasses high-readiness anchors at structural boundaries; 
    in contrast, \ours{} tends to first capture global anchors across the full sequence for initial decoding, 
    and subsequently identifies critical local tokens within each sub-segment to propagate context outward.}
    \vspace{-6pt}
  \label{fig:decoding_order}
\end{figure*}

\paragraph{Dynamic Context Score.}
Let $\mathcal{D} = \{j : x_j \neq \mask\}$ be decoded positions.
Nearest-neighbor distances to decoded left and right neighbors:
\begin{align}
  d_\ell(i) &= i - \max\{j \in \mathcal{D} : j < i\},
  \label{eq:dleft}\\
  d_r(i)    &= \min\{j \in \mathcal{D} : j > i\} - i,
  \label{eq:dright}
\end{align}
with $d_\ell(i) = d_r(i) = \infty$ when no neighbor exists.
The context score aggregates both via harmonic decay:
\begin{equation}
  \operatorname{ctx}(i)
  = \frac{1}{1+d_\ell(i)} + \frac{1}{1+d_r(i)},
  \quad \operatorname{ctx}(i) \in [0,2].
  \label{eq:ctx}
\end{equation}

\paragraph{Composite Score.}
The composite score treats token confidence as a base weight and the
context score as a multiplicative amplifier:
\begin{equation}
  \operatorname{score}(i) = c_i \cdot \bigl(1 + \beta \cdot \operatorname{ctx}(i)\bigr).
  \label{eq:score}
\end{equation}

The multiplicative form ensures contextual readiness scales
proportionally with confidence: a high-confidence token adjacent to
decoded neighbors receives a strong boost, while a low-confidence
token gains only a modest increment, suppressing noisy predictions
from being promoted on position alone.
Crucially, a low-confidence token with strong bilateral support is
still promoted above an equally confident isolated token, rescuing
contextually ready candidates from the ``hard zone''
(Figure~\ref{fig:motivation_analysis}b).
When no decoded neighbor exists, $\operatorname{ctx}(i){=}0$ and the
amplifier collapses to unity, reducing the score to pure confidence
$c_i$ and ensuring isolated positions remain rankable.
$\beta{=}0$ recovers pure confidence scoring as a special case;
Appendix~\ref{app:hyper} confirms that $\beta{=}1.0$ strikes a robust 
balance between confidence and contextual influence across models.

\subsection{Full-Sequence Cluster Commitment}
\label{sec:global}

To keep high-readiness anchors globally accessible, \ours{} operates
on the full sequence at every step: each round consumes 1~NFE, and
all positions with $\operatorname{score}(i) \geq \tau_t$ are
committed as a cluster, immediately becoming anchors that reduce
prediction difficulty for their neighbors.
Figure~\ref{fig:decoding_order} illustrates the resulting decoding
order on a representative captioning sample. Crucially, comprehensive empirical evaluations in Appendix~\ref{app:hyper} show that \ours{} remains remarkably stable across a broad range of joint $\beta$ and $\tau$ configurations, confirming that its performance gains are driven by its structural design rather than fragile hyperparameter selection.

\subsection{Architecture-Aware Calibration}
\label{sec:calibration}

The evaluated models exhibit substantially different token confidence
distributions. MMaDA has a mean confidence of 0.63, LaViDa has a
bimodal distribution with a mean of 0.47, and LLaDA-V has a strongly
left skewed distribution with a mean of 0.12, with more than 73\% of
tokens below $c_i{=}0.1$. Notably, LaViDa and LLaDA-V share the same
SigLIP encoder but produce markedly different distributions. These
observations motivate architecture specific calibration, while the
complete distributions are provided in
Appendix~\ref{app:conf_dist}.

\paragraph{Dynamic threshold scheduling.}
Inspired by the two-phase commitment risk pattern in
Prophet~\cite{prophet}, we employ a dynamic scheduling mechanism
to shift from strict precision to completion. Early phases favor
high thresholds to protect anchor quality and ensure initial
accuracy, while later phases use lower thresholds to prevent
deadlock:
\begin{equation}
  \tau_t =
  \begin{cases}
    \tau & \text{if } r_t > \delta, \\[4pt]
    \tau - \dfrac{\delta - r_t}{\delta}(\tau - \tau_{\min})
          & \text{otherwise,}
  \end{cases}
  \label{eq:tau_schedule}
\end{equation}
where $r_t = N_{\text{mask}}^t / L$ and $\tau_{\min} = 0.75\,\tau$.

\begin{algorithm}[H]
\caption{\ours{}: Context-Aware Cluster Decoding (one round)}
\label{alg:clusterdec}
\begin{algorithmic}[1]
\REQUIRE Sequence $\mathbf{x}$; parameters $\tau, \beta, \delta, \rho$; EOS token id
\STATE $p \leftarrow N_{\text{dec,non-EOS}}^t / L$;\quad
       $\epsilon \leftarrow \max(0,\, 1 - p/\rho)$
\STATE $r_t \leftarrow N_{\text{mask}}^t / L$;\quad
       $\tau_t \leftarrow \textsc{Schedule}(\tau, r_t, \delta)$
       \hfill \algcomment{Eq.~\ref{eq:tau_schedule}}
\STATE $\mathbf{l} \leftarrow f_\theta(\mathbf{x})$
  \hfill \algcomment{full-sequence forward pass, 1 NFE}
\STATE Compute $c_i$, $\operatorname{ctx}(i)$, $\operatorname{score}(i)$ for all masked $i$
  \hfill \algcomment{Eqs.~\ref{eq:conf_def}--\ref{eq:score}}
\STATE $\operatorname{score}_{\text{EOS}} \leftarrow \operatorname{score}_{\text{EOS}} \cdot (1 - \epsilon)$
  \hfill \algcomment{suppress premature EOS}
\STATE $\mathcal{C} \leftarrow \{i : \operatorname{score}(i) \geq \tau_t,\; x_i = \mask\}$
\IF{$\mathcal{C} = \emptyset$}
  \STATE $\mathcal{C} \leftarrow \{\arg\max_i\,\operatorname{score}(i)\}$
  \hfill \algcomment{deadlock prevention}
\ENDIF
\STATE $x_i \leftarrow \hat{x}_i$ \textbf{ for all } $i \in \mathcal{C}$
\end{algorithmic}
\end{algorithm}

\textbf{EOS soft suppression.}
EOS tokens accumulate spuriously high confidence before generation is
complete, biasing the model toward short outputs.
We suppress this by applying a progress-dependent penalty:
\begin{equation}
  \epsilon = \max\!\left(0,\; 1 - \frac{p}{\rho}\right),
  \label{eq:eos_epsilon}
\end{equation}
\begin{equation}
  \operatorname{score}_{\text{EOS}}(i) \leftarrow
  \operatorname{score}_{\text{EOS}}(i) \cdot (1-\epsilon),
  \label{eq:eos_suppress}
\end{equation}
where $p = N_{\text{dec,non-EOS}}^t / L$ is the fraction of
non-EOS tokens decoded so far. For $p \geq \rho$ the penalty
vanishes and EOS competes freely.

\begin{table*}[t]
\centering
\caption{Generation quality across three dMLLMs (within-model comparisons).
  \textbf{Bold}: best per model per metric. \underline{Underline}: second best.
  LLaVABench-COCO for long-form generation and CHAIR for hallucination mitigation.}
\label{tab:main}
\small
\setlength{\tabcolsep}{6pt}
\renewcommand{\arraystretch}{1.2}
\begin{tabular}{@{}ll cccccc@{}}
\toprule
\multirow{2}{*}{\textbf{Model}}
  & \multirow{2}{*}{\textbf{Method}}
  & \multicolumn{3}{c}{\textbf{Understanding \& Reasoning}}
  & \multicolumn{3}{c}{\textbf{Hallucination}} \\
\cmidrule(lr){3-5}\cmidrule(lr){6-8}
  &
  & \textbf{MathVista}$\uparrow$
  & \textbf{LLaVABench-COCO}$\uparrow$
  & \textbf{ScienceQA}$\uparrow$
  & \textbf{CHAIR\textsubscript{s}}$\downarrow$
  & \textbf{CHAIR\textsubscript{i}}$\downarrow$
  & \textbf{Recall}$\uparrow$ \\
\midrule
\rowcolor[gray]{0.92}
\multicolumn{8}{l}{\textit{\textbf{MMaDA-8B-Base}}} \\
  & Original
    & 23.50 & 35.30 & 50.24
    & 24.58 & 11.27 & 39.46 \\
  & AdaBlock
    & \underline{24.80} & \underline{47.40} & \underline{51.61}
    & 22.84 & 11.27 & \textbf{41.83} \\
  & Wavefront
    & 23.00 & 42.50 & 50.77
    & \textbf{18.62} & \textbf{9.43}  & 38.91 \\
  & \ours{}
    & \textbf{28.00} & \textbf{49.00} & \textbf{51.71}
    & \underline{19.76} & \underline{10.28} & \underline{40.63} \\
\midrule
\rowcolor[gray]{0.92}
\multicolumn{8}{l}{\textit{\textbf{LaViDa (SigLIP+LLaDA-8B)}}} \\
  & Original
    & 41.50 & 51.10 & 71.24
    & 6.93  & 10.20 & \underline{34.68} \\
  & AdaBlock
    & \underline{45.50} & 78.60 & \underline{72.93}
    & 6.62  & \underline{5.02}  & 31.62 \\
  & Wavefront
    & 45.30 & \underline{82.10} & 72.09
    & \underline{5.87}  & \textbf{4.31}  & 31.72 \\
  & \ours{}
    & \textbf{45.70} & \textbf{88.70} & \textbf{73.03}
    & \textbf{5.08} & 7.58 & \textbf{35.31} \\
\midrule
\rowcolor[gray]{0.92}
\multicolumn{8}{l}{\textit{\textbf{LLaDA-V (SigLIP+LLaDA-8B)}}} \\
  & Original
    & 30.60 & 50.30 & \underline{79.23}
    & 5.60  & \underline{3.95}  & 36.55 \\
  & AdaBlock
    & 31.70 & \underline{72.90} & 79.18
    & 5.82  & 4.17  & \underline{37.05} \\
  & Wavefront
    & \textbf{33.50} & 71.30 & 77.84
    & \textbf{5.21}  & 4.17  & 30.06 \\
  & \ours{}
    & \textbf{33.50} & \textbf{73.90} & \textbf{79.52}
    & \underline{5.38} & \textbf{3.81} & \textbf{37.30} \\
\bottomrule
\end{tabular}
\end{table*}

By committing semantically anchored tokens first and propagating
their local context outward, \ours{} helps each position decode
with accurate local semantic support from its neighborhood.

\section{Experiments}
\label{sec:exp}

\subsection{Experimental Setup}
\label{sec:setup}

We evaluate on three dMLLMs: \textbf{MMaDA-8B-Base}~\cite{mmada},
\textbf{LaViDa}~\cite{lavida}, and \textbf{LLaDA-V}~\cite{lladav}.
We compare against three baselines: \textit{Original} (standard
fixed-step block-diffusion), \textit{AdaBlock}~\cite{adablock}
(ICLR~2026), and \textit{Wavefront}~\cite{wavefront} (ICLR~2026),
originally proposed for text-only dLLMs, we evaluate them on a broad range of multimodal benchmarks. Benchmarks cover long-form generation and hallucination via \textit{LLaVABench-COCO}~\cite{LLaVA} (Relative Score) and \textit{CHAIR}~\cite{chair}, while we evaluate reasoning and classification performance via \textit{MathVista}~\cite{lu2024mathvista} (LLM-judge accuracy) as well as \textit{ScienceQA}~\cite{DBLP:journals/corr/abs-2209-09513}.

\subsection{Main Results}
\label{sec:main}

Table~\ref{tab:main} reports generation quality across all models and
benchmarks.

\textbf{Long-form generation quality.}
\ours{} achieves the highest LLaVABench-COCO score across all three
models: on LaViDa, \ours{} improves from 51.1 (Original) to 88.7
(+37.6); on LLaDA-V, from 50.3 to 73.9 (+23.6); on MMaDA, from
35.3 to 49.0 (+13.7). These gains confirm that neighbor-aware
commitment consistently benefits long-form generation regardless of
the underlying dMLLM architecture.

\textbf{Hallucination reduction.}
Relative to Original, \ours{} reduces both CHAIR metrics on MMaDA:
CHAIR$_\text{s}$ drops from 24.58 to 19.76 ($-4.82$), and
CHAIR$_\text{i}$ from 11.27 to 10.28 ($-0.99$), although Wavefront
performs better on both metrics. On LLaDA-V, \ours{} obtains the
lowest CHAIR$_\text{i}$ (3.81) and the highest Recall (37.30),
showing a favorable balance between object hallucination and coverage.
Qualitative examples in Appendix~\ref{app:casestudy} further
illustrate that \ours{} can produce descriptions with fewer
hallucinated objects while preserving relevant visual content.

\begin{table*}[!t]
\centering
\caption{Generation quality on LLaDA-8B-Instruct.
All values are percentages. \textbf{Bold}: better result.}
\label{tab:text_only_quality}

\begin{tabular}{llrrr}
\toprule
\textbf{Benchmark} & \textbf{Metric}
  & \textbf{Original}
  & \textbf{\ours{}}
  & $\boldsymbol{\Delta}$ \\
\midrule
GSM8K
  & Accuracy
  & 61.00 & \textbf{63.00} & +2.00 \\
\midrule
\multirow{4}{*}{IFEval}
  & Prompt strict
  & 59.00 & \textbf{60.00} & +1.00 \\
  & Instruction strict
  & 64.74 & \textbf{68.59} & +3.85 \\
  & Prompt loose
  & 60.00 & \textbf{61.00} & +1.00 \\
  & Instruction loose
  & 67.95 & \textbf{69.23} & +1.28 \\
\midrule
\multirow{2}{*}{AlpacaEval}
  & Win rate
  & 4.50 & \textbf{6.00} & +1.50 \\
  & Length controlled win rate
  & 10.95 & \textbf{11.45} & +0.50 \\
\bottomrule
\end{tabular}
\end{table*}

\textbf{Reasoning and classification.}
\ours{} achieves the highest MathVista accuracy on MMaDA
(from 23.5\% to 28.0\%, +4.5\%) and on LLaDA-V (from 30.6\% to
33.5\%, +2.9\%), and delivers the best ScienceQA accuracy across all
three models. Even on short outputs where void accumulation is
minimal, committing the highest-readiness token first establishes a
more reliable semantic foundation for subsequent positions, yielding
modest but consistent quality gains across benchmarks.

\textbf{Baseline comparison.}
AdaBlock and Wavefront underperform Original on several combinations
(e.g., Wavefront on MMaDA MathVista: 23.0\% vs.\ 23.5\%), particularly
on long-form tasks. As analyzed below, this degradation stems from 
a clear mismatch: existing text-only decoding assumptions fail to 
account for the unique confidence landscapes shaped by visual integration.

\textbf{Why existing decoding methods degrade on dMLLMs.}
Different visual integration strategies shape dMLLM confidence 
landscapes in ways text-only methods overlook (see Figure~\ref{fig:conf_dist}):
MMaDA's VQ-tokenizer yields a moderate token confidence distribution 
($\mu=0.63$), whereas LLaDA-V's SigLIP integration produces a heavily 
left-skewed profile ($\mu=0.12$) with over 73\% of tokens falling 
below $c_i=0.1$.

\subsection{Extension to Text Only dLLMs}
\label{sec:text_only}

Although \ours{} is designed and primarily evaluated for multimodal
diffusion models, its commitment rule depends only on token
confidence and the positions of decoded neighbors. It does not
require visual features or a model specific multimodal component.
We therefore apply \ours{} to LLaDA-8B-Instruct and evaluate it on
GSM8K, IFEval, and AlpacaEval.

As shown in Table~\ref{tab:text_only_quality}, \ours{} improves all
reported quality metrics over Original. Accuracy on GSM8K increases
from 61.00 to 63.00. On IFEval, the largest improvement is observed
for instruction level strict accuracy, which increases from 64.74
to 68.59. On AlpacaEval, the standard and length controlled win
rates increase from 4.50 to 6.00 and from 10.95 to 11.45,
respectively. These results provide preliminary evidence that
context aware commitment can also benefit text generation.

\noindent\textbf{Decoding efficiency.}
Original uses 128 fixed forward evaluations, whereas \ours{}
dynamically determines the number of commitment rounds. Average NFE
decreases to 48.00 on GSM8K, 112.73 on IFEval, and 121.75 on
AlpacaEval, producing measured speedups of $2.68\times$,
$1.20\times$, and $1.03\times$, respectively. Thus, \ours{} reduces
NFE on all three benchmarks, although the corresponding runtime gain
is modest on AlpacaEval. Complete efficiency results are provided in
Appendix~\ref{app:text_only_efficiency}.

\subsection{Quality Scales with Sequence Length}
\label{sec:seqlen}
Table~\ref{tab:seqlen} tests whether the quality advantage of \ours{}
grows with output length. At 16 tokens, Original and \ours{} are
close (45.9 vs.\ 52.2). As sequence length increases, the quality
gap widens monotonically, reaching a 44.8\% relative improvement at
1024 tokens (47.8 vs.\ 33.0) while Original continues to degrade,
confirming that \ours{} specifically addresses the long-form quality
degradation that block-based approaches increasingly suffer as
generation extends.
\begin{table}[t]
\centering
\caption{LLaVABench-COCO score and average output length.
Output length alone does not account for the observed quality
differences. \textbf{Bold}: best score per model.}
\label{tab:seqlen}
\Large
\renewcommand{\arraystretch}{1.2}
\resizebox{\linewidth}{!}{
\begin{tabular}{ccccc}
\toprule
\rowcolor[gray]{0.92}
{\textbf{Length}} & {\textbf{Original}} & {\textbf{AdaBlock}}
  & {\textbf{Wavefront}} & {\textbf{CACD}} \\
\midrule
16   & 45.9 & \textbf{52.6} & 41.9 & 52.2 \\
32   & 40.8 & \textbf{61.7} & 45.5 & 59.4 \\
64   & 33.5 & 55.9 & 46.4 & \textbf{58.8} \\
128  & 31.6 & 46.8 & 43.0 & \textbf{50.4} \\
256  & 32.3 & 46.0 & 44.6 & \textbf{47.3} \\
512  & 35.3 & 47.4 & 42.5 & \textbf{49.0} \\
1024 & 33.0 & 45.5 & 43.2 & \textbf{47.8} \\
\bottomrule
\end{tabular}}
\end{table}

\begin{table*}[t]
\centering
\caption{LLaVABench-COCO score and average output length.
  Higher score with shorter output confirms quality gains are
  not length-driven. \textbf{Bold}: best Score per model.}
\label{tab:length_score}

\begin{tabular}{@{}lcccccc@{}}
\toprule
\multirow{2}{*}{\textbf{Method}}
  & \multicolumn{2}{c}{\textbf{MMaDA}}
  & \multicolumn{2}{c}{\textbf{LaViDa}}
  & \multicolumn{2}{c}{\textbf{LLaDA-V}} \\ \cmidrule(lr){2-3} \cmidrule(lr){4-5} \cmidrule(lr){6-7}
  & {\textbf{Score}$\uparrow$} & {\textbf{Len}}
  & {\textbf{Score}$\uparrow$} & {\textbf{Len}}
  & {\textbf{Score}$\uparrow$} & {\textbf{Len}} \\
\midrule
Original  & 35.3 & 188.8 & 51.1 & 101.5 & 50.3 & 392.5 \\
AdaBlock  & 47.4 & 160.2 & 78.6 & 256.1 & 72.9 & 409.7 \\
Wavefront & 42.5 & 402.8 & 82.1 & 481.8 & 71.3 & 400.9 \\
\ours{}
  & \textbf{49.0} & 158.1
  & \textbf{88.7} & 461.2
  & \textbf{73.9} & 367.5 \\
\bottomrule
\end{tabular}
\end{table*}

Table~\ref{tab:length_score} further confirms that \ours{}'s quality
gains are not attributable to longer outputs: Wavefront produces the
longest outputs on MMaDA (402.8 chars) yet scores lower (42.5
vs.\ 49.0), ruling out length inflation as a confound.

\begin{table}[t]
\centering
\caption{Component ablation on MMaDA LLaVABench-COCO. \textbf{Bold}: best.}
\label{tab:ablation}

\setlength{\tabcolsep}{7pt}
\begin{tabular}{cccc}
\toprule
\rowcolor[gray]{0.92}
{\textbf{Ctx}} & {\textbf{Dyn.\,$\tau$}} & {\textbf{EOS}}
  & {\textbf{Score}$\uparrow$} \\
\midrule
\ding{55} & \ding{51} & \ding{51} & 46.4 \\
\ding{51} & \ding{55} & \ding{51} & 46.7 \\
\ding{51} & \ding{51} & \ding{55} & 48.2 \\
\ding{51} & \ding{51} & \ding{51} & \textbf{49.0} \\
\bottomrule
\end{tabular}
\end{table}

\subsection{Ablation Study}
\label{sec:ablation}

Table~\ref{tab:ablation} evaluates each component of \ours{} on
MMaDA LLaVABench-COCO. Additional sensitivity results are provided
in Appendix~\ref{app:hyper}. In these experiments, the context weight
$\beta{=}1.0$ performs consistently well across all three
architectures despite their different visual integration strategies.

Removing the context score (\textit{w/o Ctx}) lowers the score from
49.0 to 46.4. In this setting, the composite score reduces to pure
confidence, so tokens with stronger contextual support may remain
deferred. Replacing the dynamic threshold schedule with a fixed
threshold (\textit{w/o Dyn.\,$\tau$}) reduces the score to 46.7,
reflecting the difficulty of balancing reliable early commitments
with completion in later rounds. Without EOS suppression
(\textit{w/o EOS}), the score decreases to 48.2 because EOS may be
committed before sufficient semantic content has been generated.
Overall, each component provides a complementary contribution to
the full method.

\begin{figure}[!t]
  \centering
  \includegraphics[width=\columnwidth]{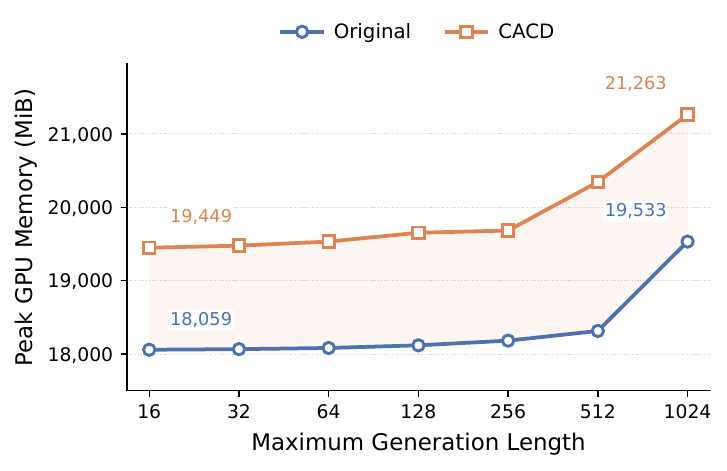}
  \vspace{-4pt}
  \caption{Peak GPU memory across maximum generation lengths on
  MMaDA, measured on an NVIDIA A800. Both methods show gradual
  memory growth, while \ours{} uses moderately more memory than
  Original across the evaluated lengths.}
  \label{fig:length_memory_scaling}
  \vspace{-5pt}
\end{figure}

\noindent\textbf{Computational cost.}
\ours{} achieves the lowest NFE on all three evaluated models.
Figure~\ref{fig:length_memory_scaling} shows how peak GPU memory
changes with the maximum generation length on MMaDA. Memory usage
increases gradually for both methods, with \ours{} requiring
approximately 1.4 to 2.0 GiB more memory than Original while
substantially reducing the number of decoding rounds. Complete NFE,
runtime, and memory comparisons across models are provided in
Appendix~\ref{app:efficiency}.

\section{Conclusion}
\label{sec:conclusion}

We presented \ours{}, a training-free decoding method that improves
long-form generation quality in dMLLMs via a multiplicative
confidence-context score that promotes contextually ready tokens
regardless of their absolute confidence level, committing them as
a cluster while operating block-free to keep high-readiness anchors
globally accessible.
Experiments across three dMLLMs and four benchmarks show consistent
quality improvement and hallucination reduction, with the advantage
compounding as outputs grow longer, confirming the void-compounding
hypothesis that misdeferred tokens cascadingly degrade local semantic
support downstream. A key finding is that visual integration
strategies shape the confidence landscape in ways existing text-only
decoding methods do not account for, a dimension that warrants
explicit treatment in future dMLLM decoding method design.

\section{Limitations}
\label{sec:limitations}

\ours{} has several practical limitations. Per-model threshold
calibration is currently required because the optimal $\tau$ varies
with each architecture's confidence distribution; automatic
calibration remains an important future direction. Although \ours{}
reduces NFE, its runtime and memory benefits are model dependent,
and integration with effective caching mechanisms may further improve
overall efficiency. Finally, our experiments demonstrate
cross-architecture generalization at the approximately 8B scale, but
not cross-scale generalization, as larger dense dMLLMs are not
currently publicly available. Evaluation at larger model scales is
therefore left for future work.

\bibliography{custom}

\appendix

\section{Extended Related Work}
\label{app:related}
\paragraph{Masked Diffusion Foundations and the Broader dLLM Ecosystem.}
The probabilistic backbone of modern dLLMs is established by
foundational works on discrete and masked diffusion, including
SEDD~\cite{sedd}, MD4~\cite{md4}, MDLM~\cite{mdlm}, RADD~\cite{radd},
and SMDM~\cite{smdm}, which together develop the training objectives,
scaling recipes, and theoretical properties that 8B+ models such as
LLaDA build upon. At the frontier, Dream-7B~\cite{dream7b},
LLaDA2.0~\cite{llada2}, and SeedDiffusion~\cite{SeedDiffusion} scale
masked diffusion to larger parameter counts, while
DiffusionVL~\cite{diffusionvl} and Dream-VL~\cite{dreamvl} extend it
to vision-language tasks beyond the three dMLLMs we evaluate.
Surveys by~\citet{surveydisc} and~\citet{openproblems} map the full
landscape, with open challenges around long-form coherence and
decoding quality directly motivating this work.

\paragraph{Decoding Structure: Block Variants and Commitment Strategies.}
Beyond the methods discussed in the main text, Swordsman~\cite{swordsman}
uses entropy signals to set block size adaptively; R3~\cite{r3block}
applies a review-remask-refine cycle within blocks; and
D2F~\cite{d2f} introduces block-wise autoregressive generation to
enable KV-cache reuse with inter-block parallel decoding.
On the commitment-order side, COVER~\cite{cover}, Parallel
Sampling~\cite{parallelsamp}, and Finish First, Perfect
Later~\cite{finishfirst} propose distinct criteria for token
commitment. Zhao and Cai~\cite{adaptintrinsic} show theoretically
that schedules adapting to the intrinsic dependence structure of the
target distribution yield improved convergence, grounding
dependency-aware strategies such as \ours{}.
Speculative decoding~\cite{specdiff,spiffy} offers a complementary
acceleration path without modifying commitment order.

\paragraph{Efficiency: Caching, Distillation, and Consistency.}
Several methods address KV-cache reuse under diffusion's bidirectional
attention~\cite{fastdllmv2,d2cache,focusdllm,mosaic,MaskKV}; notably,
WeDLM~\cite{wedlm} replaces bidirectional attention with causal
attention via topological reordering, making prefix KV-cache
compatible with parallel diffusion decoding.
Distillation and consistency training~\cite{distiltime,cdlm} reduce
required forward passes by training the model to denoise in fewer
steps, providing orthogonal strategies to further accelerate generation
that are fully complementary to our inference-time improvements. More
broadly, developing effective KV-cache mechanisms tailored for dMLLMs
remains an active area of research.
\section{Decoding Order Comparison}
\label{app:decoding_order}
\begin{figure*}[t]
  \centering
  \includegraphics[width=\textwidth]{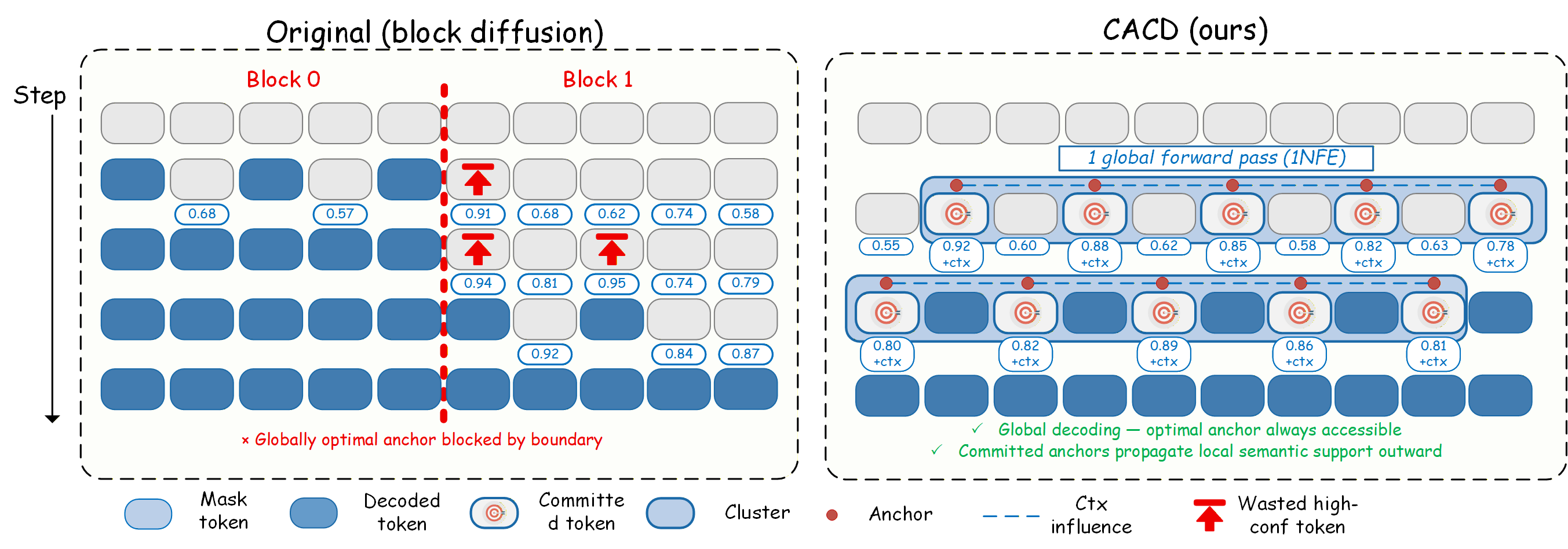}
  \caption{Decoding order comparison: block diffusion (left) vs.\ \ours{} (right).
    Block diffusion commits tokens within rigid sequential blocks, leaving
    the high-readiness anchor structurally inaccessible until its specific block is reached.
    \ours{} performs a global forward pass each round, committing
    the highest-readiness tokens as a dense cluster with context-amplified scores.}
  \label{fig:decoding_comparison}
\end{figure*}

Figure~\ref{fig:decoding_comparison} contrasts the two decoding
strategies on a representative sample.
In block diffusion (left), the token with high confidence (0.95)
is structurally blocked by the sequential block boundary: lower-confidence
tokens within the current active block are forced to commit first,
instantiating the \textit{void-compounding} effect described in
Section~\ref{sec:mot_semantic}.
By contrast, \ours{} (right) operates on the full sequence each round;
committed anchors immediately propagate context influence to neighboring
masked positions, boosting their composite scores (visualized as
``+ctx'' increments), and building a dense, locally supported
commitment front that prevents cascading generation errors.

\section{Confidence Distribution Analysis}
\label{app:conf_dist}

\begin{figure*}[!t]
  \centering
  \includegraphics[width=\textwidth]{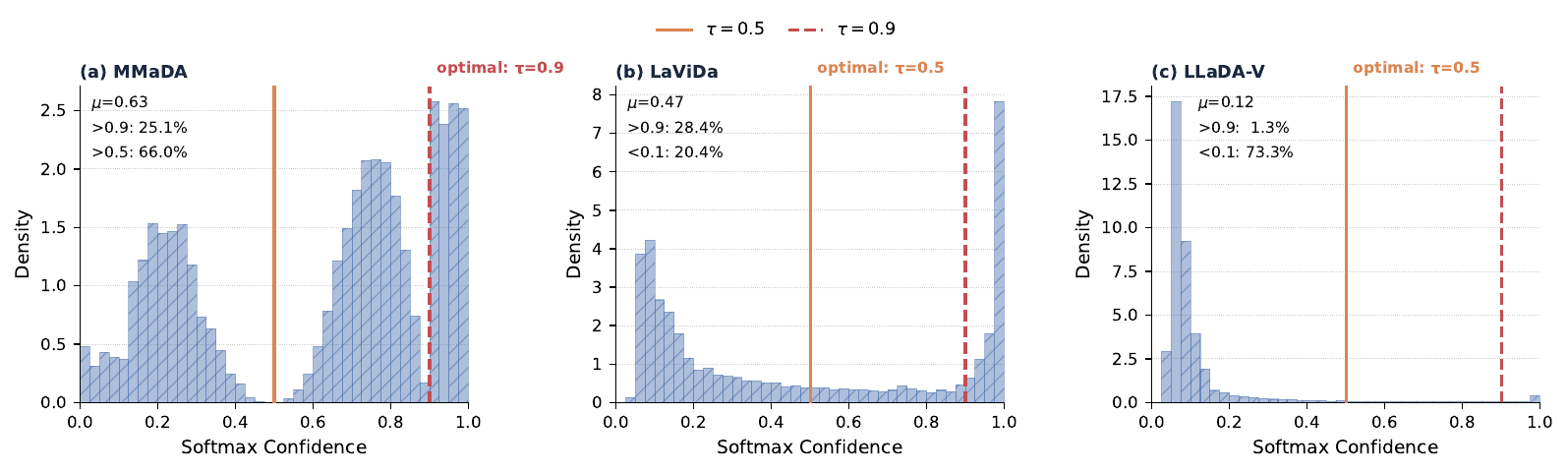}
  \vspace{-4pt}
  \caption{Token confidence distributions measured on 100 MathVista
  samples per model. MMaDA, LaViDa, and LLaDA-V exhibit substantially
  different distributions. LaViDa and LLaDA-V share the same SigLIP
  encoder, suggesting that the confidence landscape also depends on
  other model and training design choices.}
  \label{fig:conf_dist}
  \vspace{-6pt}
\end{figure*}

Figure~\ref{fig:conf_dist} presents the complete confidence
distributions for the three models. MMaDA has a relatively uniform
distribution with a mean confidence of 0.63, while LaViDa exhibits
a bimodal distribution with a mean of 0.47. LLaDA-V has a strongly
left skewed distribution with a mean of 0.12, and more than 73\% of
its tokens have confidence below $c_i{=}0.1$. These differences
explain why a single raw confidence threshold does not transfer
reliably across the evaluated architectures.

\section{Hyperparameter Sensitivity}
\label{app:hyper}

\paragraph{Effect of $\beta$.}
Table~\ref{tab:ablation_beta} shows $\beta{=}0$ consistently
underperforms $\beta{=}1.0$; values $\beta{>}1.0$ yield diminishing
returns, as over-weighting neighborhood context suppresses the
contribution of a token's own confidence, which remains an essential
signal for commitment quality. We adopt $\beta{=}1.0$ as the
optimal balance between the two.

\begin{table}[t]
\centering
\caption{Effect of $\beta$ on LLaVABench-COCO quality.
  $\beta{=}0$ recovers pure confidence scoring; $\beta{=}1.0$
  provides the best context amplification across all three models.
  \textbf{Bold}: best.}
\label{tab:ablation_beta}
\small
\setlength{\tabcolsep}{10pt}
\resizebox{\linewidth}{!}{
\begin{tabular}{lcc}
\toprule
\rowcolor[gray]{0.95}
{\textbf{Model}} & {$\boldsymbol{\beta}$}
  & {\textbf{Score}$\uparrow$} \\
\midrule
\multirow{5}{*}{MMaDA (8B-Base)}
  & 0.0 & 48.2 \\
  & 0.5 & 48.7 \\
  & \textbf{1.0} & \textbf{49.0} \\
  & 1.5 & 48.9 \\
  & 2.0 & 48.6 \\
\midrule
\multirow{5}{*}{LaViDa (SigLIP+LLaDA-8B)}
  & 0.0 & 79.3 \\
  & 0.5 & 82.3 \\
  & \textbf{1.0} & \textbf{88.7} \\
  & 1.5 & 81.3 \\
  & 2.0 & 77.5 \\
\midrule
\multirow{5}{*}{LLaDA-V (SigLIP+LLaDA-8B)}
  & 0.0 & 67.1 \\
  & 0.5 & 70.6 \\
  & \textbf{1.0} & \textbf{73.9} \\
  & 1.5 & 71.7 \\
  & 2.0 & 71.3 \\
\bottomrule
\end{tabular}}
\end{table}

\begin{table}[t]
\centering
\caption{Effect of $\tau$ on MathVista accuracy.
  Architecture-specific thresholds are required: the optimal $\tau$
  differs substantially across models due to confidence distribution
  heterogeneity. \textbf{Bold}: best.}
\label{tab:ablation_thr}
\small
\setlength{\tabcolsep}{8pt}
\renewcommand{\arraystretch}{1.2}
\begin{tabular}{lcc}
\toprule
\rowcolor[gray]{0.92}
{\textbf{Model}} & {$\boldsymbol{\tau}$}
  & {\textbf{Acc}$\uparrow$} \\
\midrule
\multirow{3}{*}{MMaDA (8B-Base)}
  & 0.5 & 21.1 \\
  & 0.7 & 24.3 \\
  & \textbf{0.9} & \textbf{28.0} \\
\midrule
\multirow{3}{*}{LaViDa (SigLIP+LLaDA-8B)}
  & \textbf{0.5} & \textbf{45.7} \\
  & 0.7 & 44.6 \\
  & 0.9 & 44.5 \\
\midrule
\multirow{3}{*}{LLaDA-V (SigLIP+LLaDA-8B)}
  & \textbf{0.5} & \textbf{33.5} \\
  & 0.7 & 33.3 \\
  & 0.9 & 32.8 \\
\bottomrule
\end{tabular}
\end{table}

\begin{figure}[t]
  \centering
  \includegraphics[width=1\linewidth]{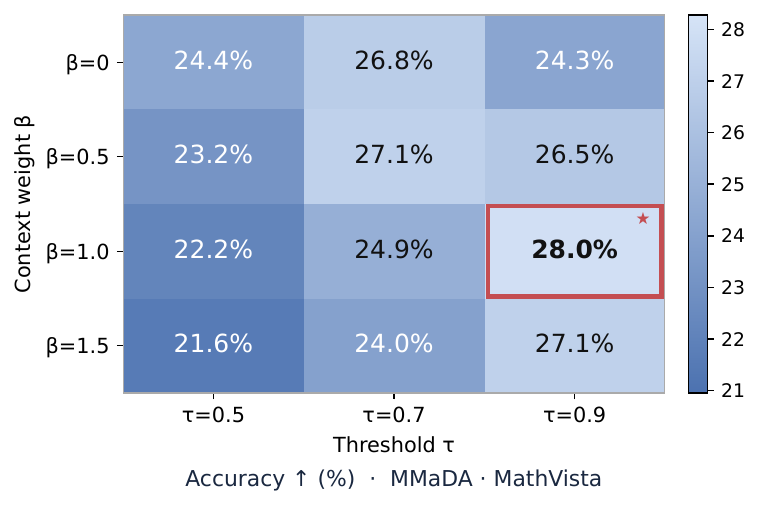}
  \vspace{-4pt}
  \caption{$\beta \times \tau$ interaction on MMaDA MathVista
    (accuracy \%). Red box: adopted configuration
    ($\beta{=}1.0$, $\tau{=}0.9$).}
    \vspace{-6pt}
  \label{fig:beta_tau}
\end{figure}

\paragraph{$\beta \times \tau$ interaction.}
Figure~\ref{fig:beta_tau} reports a grid search over
$\beta \in \{0, 0.5, 1.0, 1.5\}$ and $\tau \in \{0.5, 0.7, 0.9\}$
on MMaDA MathVista.
Accuracy increases monotonically with $\tau$ for fixed $\beta \geq
0.5$, confirming independent tuning.
Non-monotonic behavior at $\beta{=}0$ is attributable to the absence
of context weighting: without neighbor support in the scoring signal,
pure confidence alone is insufficient to maintain stable commitment
ordering, and force-commit events occasionally promote low-quality
tokens.
The optimal $\beta{=}1.0$, $\tau{=}0.9$ achieves the highest
accuracy ($28.0\%$).
This interaction also clarifies why $\beta$ generalizes across
architectures while $\tau$ does not. The context score
$\operatorname{ctx}(i) \in [0, 2]$ is a purely relative, positional
quantity independent of absolute confidence values; its amplification
effect therefore remains consistent regardless of how a model's
visual integration strategy shapes the confidence distribution.
By contrast, $\tau$ operates directly on raw confidence and is thus
sensitive to distributional shift across architectures, as confirmed
by the marked difference in optimal $\tau$ across models
(Table~\ref{tab:ablation_thr}) and the confidence heterogeneity
documented in Appendix~\ref{app:conf_dist}.

\paragraph{Effect of $\delta$.}
Too-small $\delta$ increases deadlock; too-large $\delta$ admits
low-confidence tokens early. $\delta{=}0.3$ is the universal default
(Table~\ref{tab:ablation_delta}).

\begin{table}[t]
\centering
\caption{Effect of $\delta$ on MMaDA MathVista.
  \textbf{Bold}: selected default.}
\label{tab:ablation_delta}
\small
\setlength{\tabcolsep}{15pt}
\renewcommand{\arraystretch}{1.2}
\begin{tabular}{lc}
\toprule
\rowcolor[gray]{0.92}
{\textbf{$\boldsymbol{\delta}$}} & {\textbf{Acc}$\uparrow$} \\
\midrule
0.1 & 26.9 \\
0.2 & 27.8 \\
\textbf{0.3} & \textbf{28.0} \\
0.5 & 26.8 \\
\bottomrule
\end{tabular}
\end{table}

\paragraph{Effect of $\rho$.}
Table~\ref{tab:ablation_eos} ablates the EOS release ratio $\rho$
on MMaDA LLaVABench-COCO. Score is stable across $\rho$ while
output length increases monotonically, indicating output length is
governed by the VQ-tokenizer training prior rather than $\rho$.
We adopt $\rho{=}0.8$ as the universal default.

\begin{table}[H]
\centering
\caption{Effect of EOS release ratio $\rho$ on MMaDA
  LLaVABench-COCO. \textbf{Bold}: selected default.}
\label{tab:ablation_eos}
\small
\setlength{\tabcolsep}{8pt}
\renewcommand{\arraystretch}{1.2}
\begin{tabular}{lcc}
\toprule
\rowcolor[gray]{0.92}
{\textbf{$\boldsymbol{\rho}$}} & {\textbf{Score}$\uparrow$}
  & {\textbf{Avg.\ Length (chars)}} \\
\midrule
0.5 & 47.7 & 154.9 \\
0.6 & 48.1 & 157.8 \\
\textbf{0.8} & \textbf{49.0} & 158.1 \\
1.0 & 47.8 & 161.0 \\
\bottomrule
\end{tabular}
\end{table}

\section{Decoding Cost}
\label{app:efficiency}

We compare the number of function evaluations, denoted as NFE, and
the measured decoding speed of all methods. The Original decoder uses
a fixed number of forward evaluations determined by the maximum
generation length. By contrast, \ours{} performs a full sequence
forward pass in each round and can commit multiple positions
simultaneously.

As shown in Table~\ref{tab:efficiency}, \ours{} substantially reduces
NFE on MMaDA across all three benchmarks. On LLaVABench-COCO with a
maximum generation length of 512 tokens, it reduces NFE from 512.0
to 45.9 and provides a measured speedup of $11.2\times$. This
reduction results from the cluster commitment procedure, which often
allows multiple positions to be committed within the same decoding
round.

\begin{table*}[!htbp]
\centering
\caption{Decoding cost on MMaDA-8B-Base without KV caching.
NFE denotes the number of function evaluations and Spd.\ denotes
the measured speedup. \textbf{Bold}: best.}
\label{tab:efficiency}
\small
\begin{tabular}{@{}lcccccc@{}}
\toprule
\multirow{2}{*}{\textbf{Method}}
  & \multicolumn{2}{c}{\textbf{MathVista}}
  & \multicolumn{2}{c}{\textbf{LLaVABench-COCO}}
  & \multicolumn{2}{c}{\textbf{ScienceQA}} \\
\cmidrule(lr){2-3}
\cmidrule(lr){4-5}
\cmidrule(lr){6-7}
  & \textbf{NFE}$\downarrow$ & \textbf{Spd.}$\uparrow$
  & \textbf{NFE}$\downarrow$ & \textbf{Spd.}$\uparrow$
  & \textbf{NFE}$\downarrow$ & \textbf{Spd.}$\uparrow$ \\
\midrule
Original
  & 128.0 & $1.0\times$
  & 512.0 & $1.0\times$
  & 16.00 & $1.0\times$ \\
AdaBlock
  & 25.9 & $3.9\times$
  & 74.0 & $6.5\times$
  & 1.62 & $\mathbf{6.6\times}$ \\
Wavefront
  & 60.0 & $3.8\times$
  & 128.1 & $4.0\times$
  & 4.00 & $3.4\times$ \\
\ours{}
  & \textbf{16.8} & $\mathbf{4.5\times}$
  & \textbf{45.9} & $\mathbf{11.2\times}$
  & \textbf{1.32} & $5.1\times$ \\
\bottomrule
\end{tabular}
\end{table*}

Table~\ref{tab:nfe_lavida_lladav} reports the corresponding NFE
results on LaViDa and LLaDA-V. \ours{} requires the fewest forward
evaluations on every evaluated benchmark for both models, although
the magnitude of the reduction varies across models and tasks.

\begin{table}[!htbp]
\centering
\caption{NFE on LaViDa and LLaDA-V.
Lower values are better. \textbf{Bold}: best.}
\label{tab:nfe_lavida_lladav}
\Large
\setlength{\tabcolsep}{5pt}
\renewcommand{\arraystretch}{1.12}
\resizebox{\columnwidth}{!}{
\begin{tabular}{lrrr}
\toprule
\textbf{Method}
  & \textbf{MathVista}
  & \textbf{LLaVABench-COCO}
  & \textbf{ScienceQA} \\
\midrule
\multicolumn{4}{l}{\textit{\textbf{LaViDa}}} \\
\quad Original
  & 128.0 & 512.0 & 16.0 \\
\quad AdaBlock
  & 27.4 & 52.1 & 2.1 \\
\quad Wavefront
  & 36.0 & 71.3 & 3.0 \\
\quad \ours{}
  & \textbf{16.0} & \textbf{45.6} & \textbf{1.1} \\
\midrule
\multicolumn{4}{l}{\textit{\textbf{LLaDA-V}}} \\
\quad Original
  & 128.0 & 512.0 & 16.0 \\
\quad AdaBlock
  & 45.5 & 226.6 & 7.2 \\
\quad Wavefront
  & 64.2 & 256.1 & 8.2 \\
\quad \ours{}
  & \textbf{33.1} & \textbf{164.2} & \textbf{6.5} \\
\bottomrule
\end{tabular}}
\end{table}

\subsection{Peak GPU Memory}
\label{app:peak_memory}

We additionally measure peak GPU memory on LLaVABench-COCO using
\texttt{max\_new\_tokens=512} on an NVIDIA A800 GPU.
Table~\ref{tab:peak_memory} reports NFE, measured decoding speed,
and peak memory under the same evaluation setting.

The practical tradeoffs depend on the model. On MMaDA, \ours{}
provides the largest speedup, while Original uses less peak memory.
On LaViDa, \ours{} requires the fewest forward evaluations and uses
the least peak memory, while AdaBlock is slightly faster. On
LLaDA-V, \ours{} again requires the fewest forward evaluations and
uses the least peak memory, while AdaBlock provides the largest
speedup. Thus, the NFE reduction of \ours{} does not imply uniform
superiority in runtime or memory usage.

\begin{table}[!htbp]
\centering
\caption{Decoding efficiency and peak GPU memory on
LLaVABench-COCO using \texttt{max\_new\_tokens=512} on an NVIDIA
A800. Memory is reported in MiB. The best result for each model and
metric is shown in \textbf{bold}.}
\label{tab:peak_memory}
\footnotesize
\setlength{\tabcolsep}{3.5pt}
\renewcommand{\arraystretch}{1.08}
\resizebox{\columnwidth}{!}{
\begin{tabular}{llrrr}
\toprule
\textbf{Model} & \textbf{Method}
  & \textbf{NFE}$\downarrow$
  & \textbf{Spd.}$\uparrow$
  & \textbf{Memory}$\downarrow$ \\
\midrule
\multirow{4}{*}{MMaDA}
  & Original
  & 512.0 & $1.0\times$ & \textbf{18,315} \\
  & AdaBlock
  & 74.0 & $6.5\times$ & 22,775 \\
  & Wavefront
  & 128.1 & $4.0\times$ & 21,157 \\
  & \ours{}
  & \textbf{45.9} & $\mathbf{11.2\times}$ & 20,345 \\
\midrule
\multirow{4}{*}{LaViDa}
  & Original
  & 512.0 & $1.0\times$ & 33,737 \\
  & AdaBlock
  & 52.1 & $\mathbf{2.1\times}$ & 29,205 \\
  & Wavefront
  & 71.3 & $1.3\times$ & 29,519 \\
  & \ours{}
  & \textbf{45.6} & $2.0\times$ & \textbf{28,585} \\
\midrule
\multirow{4}{*}{LLaDA-V}
  & Original
  & 512.0 & $1.0\times$ & 34,175 \\
  & AdaBlock
  & 226.6 & $\mathbf{2.6\times}$ & 36,509 \\
  & Wavefront
  & 256.1 & $1.1\times$ & 26,703 \\
  & \ours{}
  & \textbf{164.2} & $1.8\times$ & \textbf{26,529} \\
\bottomrule
\end{tabular}}
\end{table}

\section{Text Only Decoding Efficiency}
\label{app:text_only_efficiency}

Table~\ref{tab:text_only_efficiency} provides the complete decoding
efficiency results corresponding to the text generation experiments
in Section~\ref{sec:text_only}. NFE is averaged over fixed 100
example subsets under the same hardware environment. Although
\ours{} reduces NFE on all three benchmarks, the measured speedup
varies with the task and output structure.

\begin{table}[!htbp]
\centering
\caption{Decoding efficiency on LLaDA-8B-Instruct. Lower NFE and
higher speedup are better.}
\label{tab:text_only_efficiency}
\footnotesize
\setlength{\tabcolsep}{5pt}
\renewcommand{\arraystretch}{1.08}
\resizebox{\columnwidth}{!}{
\begin{tabular}{lrrr}
\toprule
\textbf{Benchmark}
  & \textbf{Original NFE}
  & \textbf{\ours{} NFE}
  & \textbf{Speedup} \\
\midrule
GSM8K
  & 128.00 & \textbf{48.00} & $\mathbf{2.68\times}$ \\
IFEval
  & 128.00 & \textbf{112.73} & $\mathbf{1.20\times}$ \\
AlpacaEval
  & 128.00 & \textbf{121.75} & $\mathbf{1.03\times}$ \\
\bottomrule
\end{tabular}}
\end{table}

\section{Qualitative Case Study}
\label{app:casestudy}

\begin{figure*}[t]
  \centering
  \includegraphics[width=\textwidth]{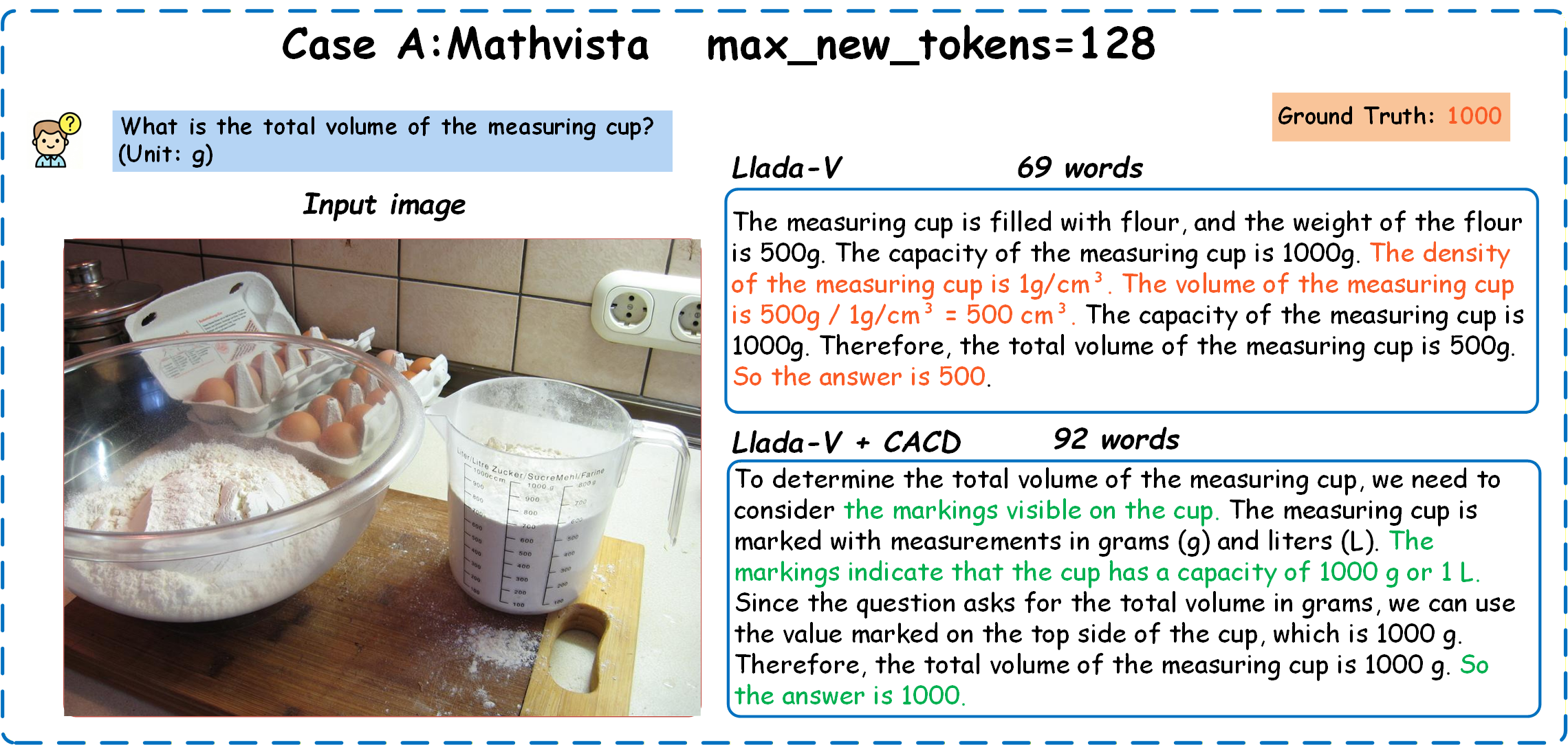}\\[6pt]
  \includegraphics[width=\textwidth]{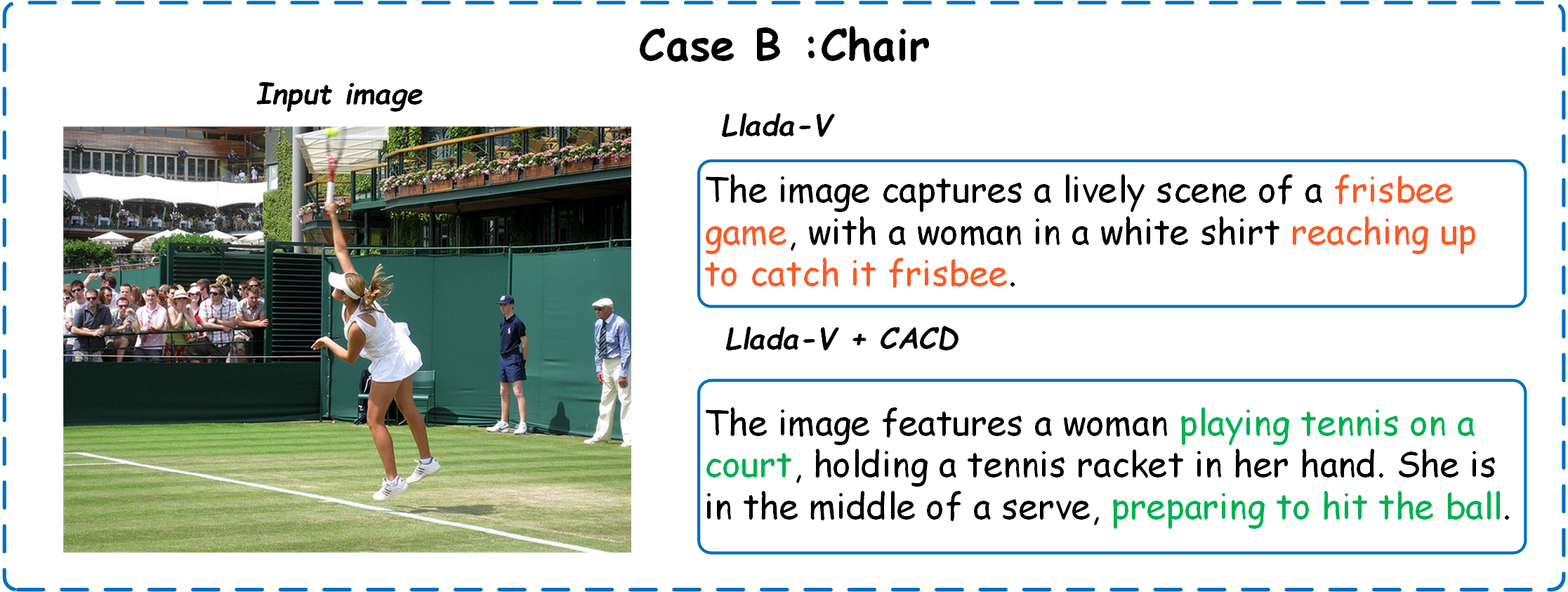}\\[6pt]
  \includegraphics[width=\textwidth]{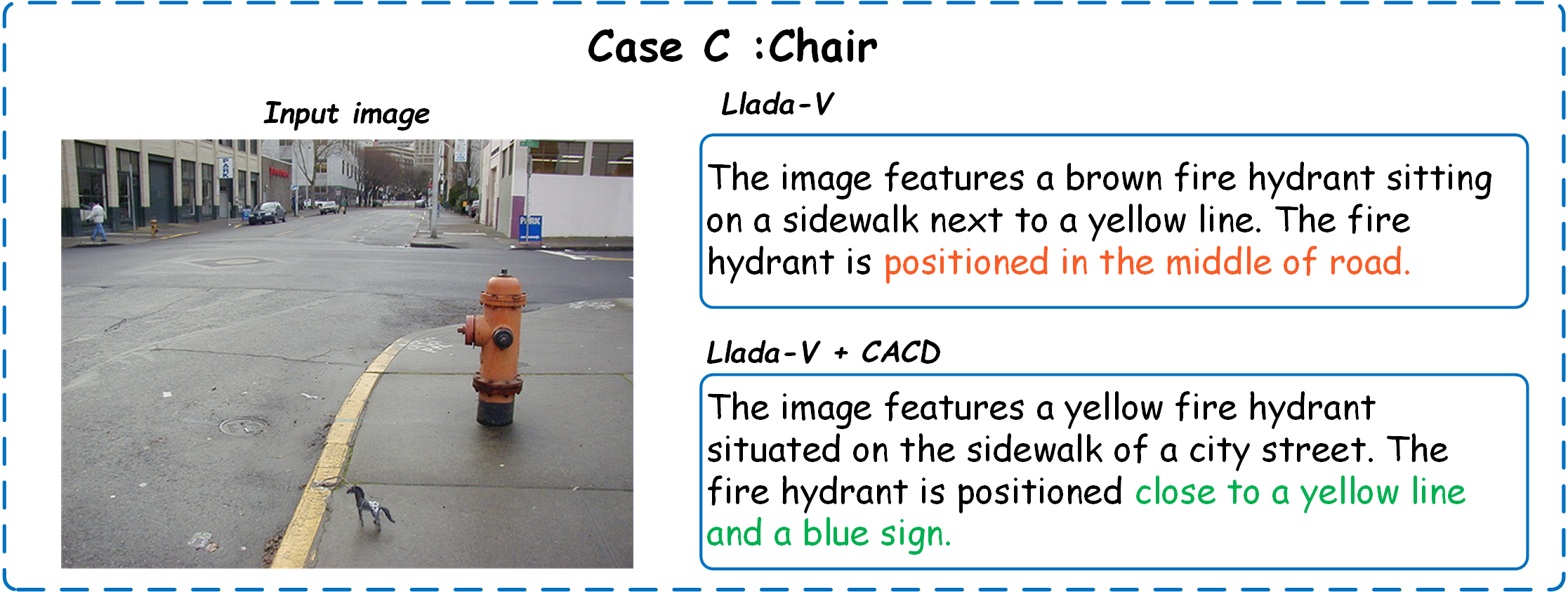}
  \caption{
    \textbf{Case A: MathVista --- Measuring Cup.}
    \textit{LLaDA-V}: reasoning drift produces the wrong answer (500).
    \textit{LLaDA-V + \ours{}}: anchors the visual token (``1000g/1L'') as the first committed cluster and propagates context outward, yielding the correct answer (1000).
    \textbf{Case B: CHAIR --- Tennis Court.}
    \textit{LLaDA-V}: object hallucination produces an incorrect scene description (``frisbee game'').
    \textit{LLaDA-V + \ours{}}: anchors the visual token (``tennis court / racket'') and propagates context outward, yielding the correct description (woman playing tennis, preparing to serve).
    \textbf{Case C: CHAIR --- Fire Hydrant.}
    \textit{LLaDA-V}: attribute hallucination produces an incorrect spatial description (``positioned in the middle of road'').
    \textit{LLaDA-V + \ours{}}: anchors the visual token (``sidewalk / yellow line'') and propagates context outward, yielding the correct description (hydrant on the sidewalk, close to a yellow line and a blue sign).
  }
  \label{fig:cases_abc}
\end{figure*}

\begin{figure*}[p]
  \centering
  \includegraphics[width=\textwidth]{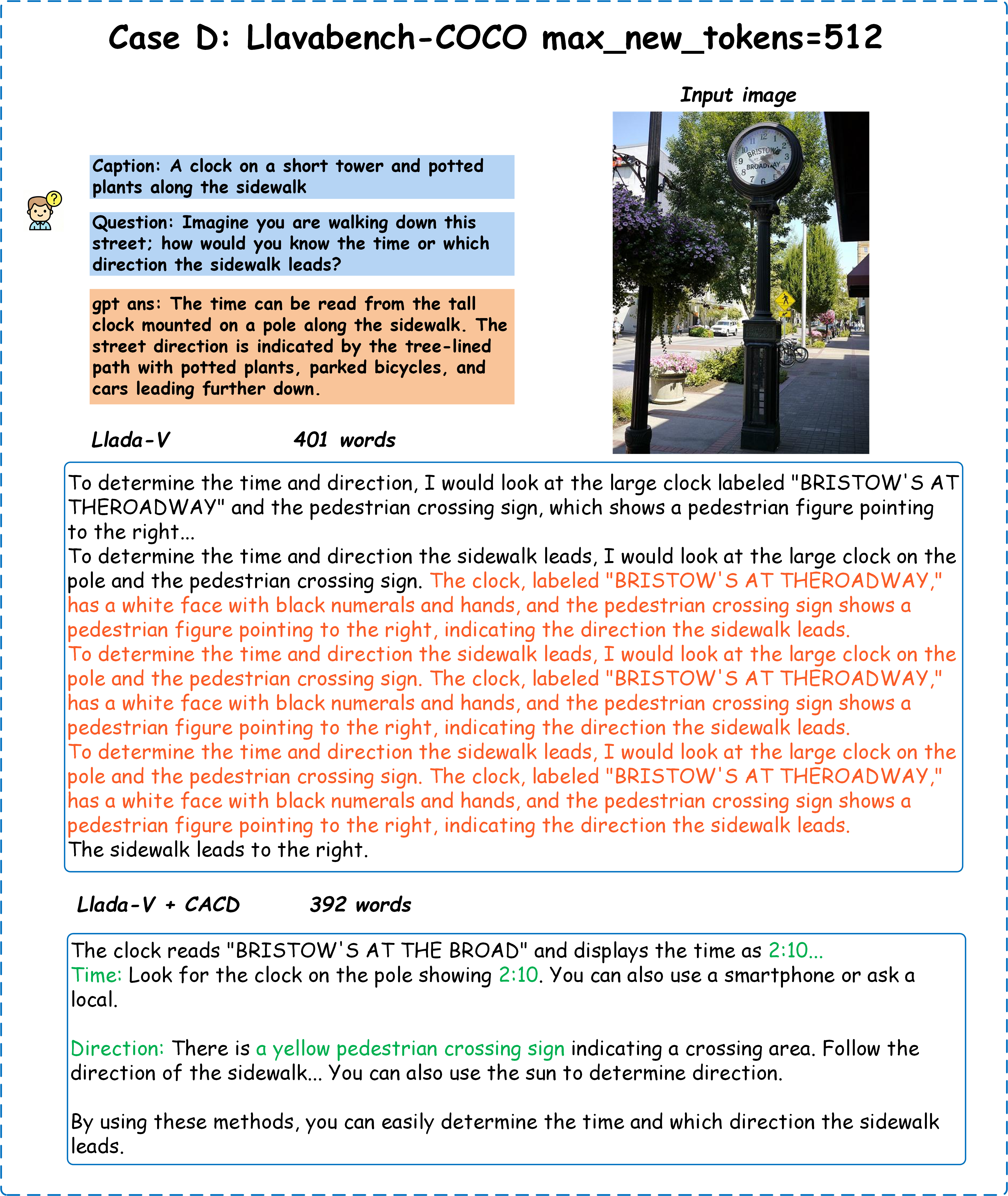}
  \caption{\textbf{Case D: LLaVABench-COCO --- Clock Tower
    (doc\_id=41).}
    \textit{Original}: repeated paragraphs and missing time readings.
    \textit{\ours{}}: well-structured response with specific readings
    and multiple candidate interpretations.}
  \label{fig:case_d}
\end{figure*}

\begin{figure*}[p]
  \centering
  \includegraphics[width=\textwidth]{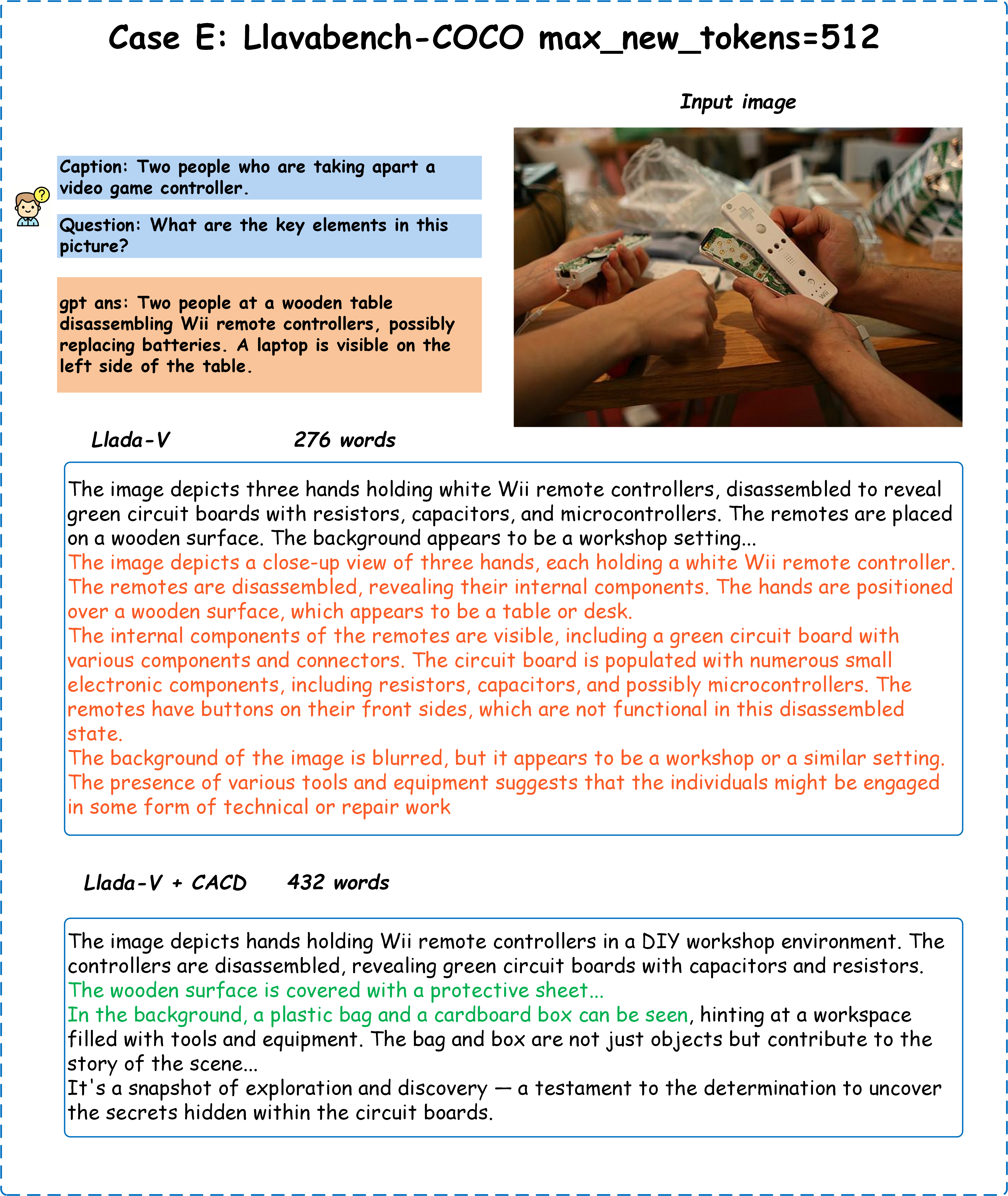}
  \caption{\textbf{Case E: LLaVABench-COCO --- Wii Controller
    (doc\_id=49).}
    \textit{Original}: verbatim repetition and truncation.
    \textit{\ours{}}: complete, non-repetitive response with broader
    scene coverage including background objects absent from the
    Original output.}
  \label{fig:case_e}
\end{figure*}

Figures~\ref{fig:cases_abc}, \ref{fig:case_d}, and~\ref{fig:case_e}
compare the Original decoding method and \ours{} on representative
MathVista, CHAIR, and LLaVABench-COCO samples.

\end{document}